%% file: main.tex
\documentclass[10pt,twocolumn,letterpaper]{article}

\usepackage[pagenumbers]{cvpr} % To force page numbers, e.g. for an arXiv version

\input{preamble}

\definecolor{cvprblue}{rgb}{0.21,0.49,0.74}
\usepackage[pagebackref,breaklinks,colorlinks,citecolor=cvprblue]{hyperref}

\def\paperID{10150} % *** Enter the Paper ID here
\def\confName{CVPR}
\def\confYear{2024}

\title{EFormer: Enhanced Transformer towards Semantic-Contour Features of Foreground for Portraits Matting}

\author{First Author\\ Zitao Wang, Qiguang Miao
\and
Second Author\\ Peipei Zhao, Yue Xi
}

\begin{document}
\maketitle
\input{sec/0_abstract}
\input{sec/1_Introduction}
% \input{sec/2_formatting}
\input{sec/2_relatedworks}
\input{sec/3_Model}
\input{sec/4_Experiments}
\input{sec/5_Conclusion}
% \input{sec/3_finalcopy}
{
    \small
    \bibliographystyle{ieeenat_fullname}
    \bibliography{main}
    % \bibliography{reference}
}

% WARNING: do not forget to delete the supplementary pages from your submission 
% \input{sec/X_suppl}

\end{document}

%% file: preamble.tex
\usepackage[dvipsnames]{xcolor}

%% file: sec/0_abstract.tex
\begin{abstract}
The portrait matting task aims to extract an alpha matte with complete semantics and finely-detailed contours. 
In comparison to CNN-based approaches, transformers with self-attention module have a better capacity to capture long-range dependencies and low-frequency semantic information of a portrait.
However, the recent research shows that self-attention mechanism struggles with modeling high-frequency contour information and capturing fine contour details, which can lead to bias while predicting the portrait's contours. 
To deal with this issue, we propose EFormer to enhance the model's attention towards both of the low-frequency semantic and high-frequency contour features. 
For the high-frequency contours, our research demonstrates that cross-attention module between different resolutions can guide our model to allocate attention appropriately to these contour regions.
Supported on this, we can successfully extract the high-frequency detail information around the portrait's contours, which are previously ignored by self-attention.
Based on cross-attention module, we further build a semantic and contour detector (SCD) to accurately capture both of the low-frequency semantic and high-frequency contour features.
And we design contour-edge extraction branch and semantic extraction branch to extract refined high-frequency contour features and complete low-frequency semantic information, respectively.
Finally, we fuse the two kinds of features and leverage segmentation head to generate a predicted portrait matte. 
Experiments on VideoMatte240K (JPEG SD Format) and Adobe Image Matting (AIM) datasets demonstrate that EFormer outperforms previous portrait matte methods.
\end{abstract}

%% file: sec/1_Introduction.tex
\section{Introduction}
\label{sec:Introduction}
% 任务介绍
Portrait matting aims to extract a precise alpha matte of persons from natural images. 
It has gathered great momentum in vision applications, including but not limited to photo editing and background replacement.

\begin{figure}[t]
  \centering
  % \fbox{\rule{0pt}{2in} \rule{0.9\linewidth}{0pt}}
   \includegraphics[width=1 \linewidth]{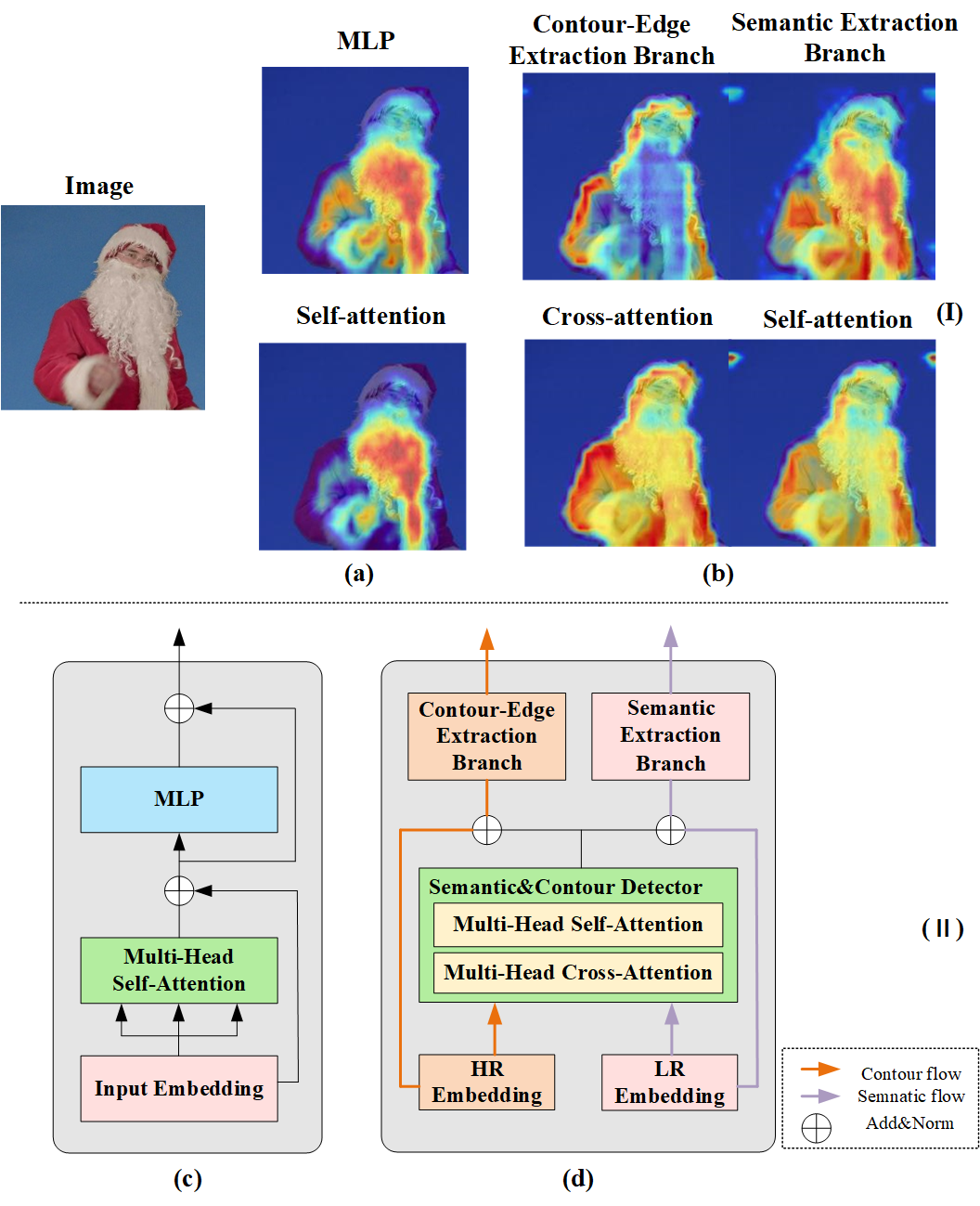}
   \caption{
    (\uppercase\expandafter{\romannumeral1}) Visualization with Grad-CAM of output from each layer within different transformer block.
    (a) The features capturing from Self-attention and MLP layers in the original transformer block.
    (b) In compared, the features capturing from each layer in the proposed transformer block.
    (\uppercase\expandafter{\romannumeral2}) Comparisons of different transformer blocks.
    (c) The original transformer block in ViT.
    (d) The proposed transformer block for portrait matte.
    }
   \label{fig:transformer-block-layer-result-compare}
\end{figure}

% CNN methods
Existing methods\cite{Such-trimap-based-approaches01,Such-trimap-based-approaches02,Such-trimap-based-approaches03,Such-trimap-based-approaches04-AIM} attempt to estimate a portrait matte with Convolutional Neural Networks (CNNs) under the guidance of a pre-determined trimap, which is a three-class map that indicates the foreground, background, and unknown region.
However, it is time-consuming and labor-intensive to obtain abundant their trimaps by manually annotate images.
Some methods eliminate the trimap, such as background matting series\cite{background-matting-series01,Other-trimap-free-solutions-background-matting-series-VideoMatte240K} uses a separate background image instead of the trimap as auxiliary reference for model prediction.
However, in practical cases, it is necessary to provide two aligned images for the model, which include one with only the background and the other one including people.
An alternative solution\cite{Other-trimap-free-solutions-MODNet-PPM-100,Other-trimap-free-solutions-RVM,Other-trimap-free-solutions04} is to rely solely on RGB images as input for predicting portrait mattes.
Although the previous models have made strides in portrait matte, there is a constraint on the receptive field in CNN-based approaches. 

% ViT models
In contrast, Vision Transformer (ViT)\cite{vision-transformer-solutions-ViT} and Swin Transformer (Swin) \cite{vision-transformer-solutions-Swin} for semantic segmentation employ a self-attention mechanism, which conducts interactions among all pixels in an image to obtain a global receptive field.
Previous research~\cite{According-to-previous-research} shows that ViT and its variants can capture the low-frequency components in images effectively, such as global shapes and structures of a scene or object. 
However, they usually neglect high-frequency ones in images, such as edges and textures. 
They work similarly to low-pass filters.

% 这里的分析挺好的，是本文的motivation
As shown in \cref{fig:transformer-block-layer-result-compare}(a), we can observe the transformer block with self-attention clearly pays little attention to the portrait's high-frequency contour regions.
We believe that this phenomenon is attributed to the excessive propagation of global information, because global attention is performed in each transformer block during the calculation of multiple blocks.
So, the global information are propagated throughout the model in this iterative calculation, ultimately leading it to only focus on low-frequency semantic features with stronger clustering in portraits.
Over time, the model gradually ignores the discrete high-frequency detail features around the contours.
Therefore, it is desired to capture low- and high-frequency components simultaneously to improve the segmentation performance.

% 其他方法改进的不足
To achieve this goal, Inception Transformer (IFormer)\cite{IFormer} splits all channels into convolutional path and self-attention path, respectively, to capture high and low frequencies. 
However, it is required for a channel ratio of each block at different levels in IFormer.
These channel ratios typically require manual adjustment, resulting in significant uncertainty when aiming for optimal model performance. 
To enhance the model’s stability and adaptive capability, we propose a novel architecture EFormer without manually adjusted parameters.
In EFormer, we use the cross-attention layer between different resolution features, as shown in \cref{fig:transformer-block-layer-result-compare}(d), which is replaced of convolution to handle high-frequency detail features.

% \begin{figure}
% \centering
% \includegraphics[width=100mm]{transformer_block_compare.png} 
% \caption{
% Comparisons of different transformer blocks.
% (a) The original transformer block in ViT.
% (b) The proposed transformer block for portrait matte.
% Our transformer block cascades cross-attention layer between features with different resolutions and self-attention layer to sequentially capture objects' contours and semantic features.
% }
% \label{fig:transformer block compare}
% \end{figure}

% 实现思路
Our research further demonstrates that the cross-attention module between different resolutions can formulate more reasonable attention allocation mechanism, which can accurately capture contour features, as shown in \cref{fig:transformer-block-layer-result-compare}(b).
Particularly, EFormer with cross-attention layer not only can not lost the semantic information within the portrait, but also can successfully capture the high-frequency details around the portrait's contours, which are previously ignored by self-attention.
It is worth noting that high-frequency components only around the boundary between foreground and background are critical to the improvement of segmentation performance.
Therefore, we follow a coarse-to-fine manner.
Initially, EFormer needs to locate all high-frequency regions in the image.
Then, it filters and extracts the high-frequency regions around the foreground contours.

% 落实细节 In detail
Based on the cross-attention module, we design a semantic and contour detector (SCD) in EFormer, which cascades cross-attention layer and self-attention layer to sequentially locate the contour and semantic features. 
Firstly, the model adjusts its attention on contour features with the guidance of cross-attention module.
Secondly, self-attention module captures semantic information within a portrait's contours, and reverses filtering and correcting the contour information output by cross-attention layer. 
This allows the model to gradually capture and match the high-frequency contour information and the low-frequency semantic information. 
Then, we use multi-layer perceptron (MLP) to build the contour-edge extraction branch (CEEB) and the semantic extraction branch (SEB), respectively. 
With this support, we independently purify the contour flow and the semantic flow to obtain finer contour information and more comprehensive portrait semantic information, as shown in \cref{fig:transformer-block-layer-result-compare}(b). 
Finally, we fuse the contour features and semantic features, and send them to the segmentation head to estimate portrait matte.
We perform extensive experiments on VideoMatte240K (JPEG SD Format)\cite{Other-trimap-free-solutions-background-matting-series-VideoMatte240K}, Adobe Image Matting (AIM)\cite{Such-trimap-based-approaches04-AIM}, and BG10K. 
Comparisons with other state-of-the-art models and ablation studies verify that our model performs better than previous works.

% detector extractor construct

% Add the contributions of the paper.
To summarize, our contributions are as follows:

\begin{itemize}

\item[$\bullet$] We propose EFormer: an approach can enhance the transformer's attention towards both semantic and contour features of foreground. 

\item[$\bullet$] We further build a semantic and contour detector (SCD) to accurately capture the semantic and contour features and design two separate extraction branches to purify the contour flow and the semantic flow.

\item[$\bullet$] We leverage the cross-attention module between different resolution features in transformer block to autonomously capture and extract high-frequency detail features, which are sparsely distributed along the foreground contour. 
\end{itemize}

% \textbullet  We propose EFormer: an approach can enhance the transformer's attention towards both semantic and contour features of foreground. 

% \textbullet  We further build a semantic and contour detector (SCD) to accurately capture the semantic and contour features and design two separate extraction branches to purify the contour stream and the semantic stream.

% \textbullet  We leverage the cross-attention module between different resolution features in transformer block to autonomously capture and extract high-frequency detail features, which are sparsely distributed along the foreground contour. 

%  Yue is here!

%% file: sec/2_relatedworks.tex
\section{Related Works}
\label{sec:Related Works}
%-------------------------------------------------------------------------
\subsection{Image Matting}
Image matting is an essential task in the field of computer vision that aims to accurately estimate the foreground in an image. Mathematically speaking, an image $I$ is a combination of an unknown foreground image $F$ and a background image $B$ with a probability coefficient alpha matte maps $\alpha$.
\begin{equation}
 I=\alpha F+\left(1-\alpha\right)B
  \label{equ:dt}
\end{equation}

Previous image matting solutions have predominantly focused on low-level features, such as color cues\cite{color-cues01,color-cues02,color-cues03,color-cues04} or propagation\cite{Propagation01,Propagation02,propagation03}, to distinguish the transition areas between foregrounds and backgrounds. 
However, such traditional matting algorithms commonly struggle to perform properly in complex scenes.

With the significant advancements making in deep learning, numerous methods based on convolutional neural networks (CNNs) are proposed, leading to notable successes. 
Some approaches\cite{Such-trimap-based-approaches05,Such-trimap-based-approaches06,Such-trimap-based-approaches07} incorporate auxiliary trimap supervisions to enhance matting performance, while other methods\cite{Other-trimap-free-solutions04,Other-trimap-free-solutions-RVM} leverage trimap-free solutions to estimate alpha mattes from image feature maps using an end-to-end segmentation network. 
In the background matting series\cite{background-matting-series01,Other-trimap-free-solutions-background-matting-series-VideoMatte240K}, an auxiliary input of the background image is used for compute alpha matte. 
Furthermore, MODNet\cite{Other-trimap-free-solutions-MODNet-PPM-100} proposes an end-to-end manner without any auxiliary input.
Although prior works based on CNN kernels have improved the accuracy of image matting, the prediction of models are in the limited receptive field.

%-------------------------------------------------------------------------
\subsection{Vision Transformer}
Different from CNNs, vision transformers with self-attention mechanisms can capture long-term dependencies.
Therefore, in the realm of computer vision, vision transformers have gained significant attention in recent times. 
For pixel level prediction tasks, such as segmentation, SETR\cite{SETR}, SegFormer\cite{SegFormer} and DPT\cite{DPT} apply transformer as encoder to attain feature maps of images.
Their performances demonstrate the transformer with self-attention is capable of building complete contextual information through global interactions.
However, recent studies\cite{According-to-previous-research,IFormer} indicate that transformer ``yet is incompetent in capturing high frequencies that predominantly convey local information''.

Compared to self-attention, cross-attention has more potential to induce customized features. 
Max-Deeplab\cite{Max-Deeplab}, MaskFormer\cite{MaskFormer01,MaskFormer02} and SeMask\cite{SeMask} use query-based methods through cross-attention, which are inspired by DETR\cite{DETR}. 
They view segmentation as a set prediction problem. 
U-Transformer\cite{U-Transformer} and EUT\cite{EUT} modify the cross-attention in the original transformer to leverage the information from the encoder, allowing a fine spatial recovery in the decoder. 
The cross-resolution attention employed by RTFormer\cite{RTFormer} enables the gathering of comprehensive contextual information for high-resolution features.

In this work, we propose a strategy that use of cross-attention between different resolutions guides the model to autonomously locate and capture high-frequency features, particularly those near the portrait's contours. 
Contrary to IFormer\cite{IFormer}, this method eliminates the need for predefined parameters to enhance the model's adaptive capabilities. 
At the same time, this approach can significantly improve the model's performance in capturing contour details.

%-------------------------------------------------------------------------
\subsection{Refinement for Segmentation}
In the field of portrait segmentation, refining the high-frequency details near the contours is crucial.
Many prior studies on segmentation refinement depend on convolutional networks or MLPs that are designed specifically for this purpose. 

PointRend\cite{PointRend}, based on the coarse predictions of Mask R-CNN\cite{Mask-R-CNN}, calculates confidence scores as the point-wise uncertainty measure. 
And it focuses on the points with the higher uncertainty according to set hyperparameters. 
Finally, PointRend uses a shared MLP to refine the labels of the selected points. 
Although PointRend has implemented segmentation refinement to some extent, the model still requires manual adjustment of certain hyperparameters to assist in filtering uncertain points. 
Transfiner\cite{Transfiner} processes detected error-prone and incoherent regions with transformer, which are including high-frequency regions strewn along object boundaries.
However, during this procedure, excessive focus is placed on the local details and incoherent regions, without effectively associating the local features of the object with the overall semantic information.

We believe that locating effective detail distribution areas requires information interaction between local details and global semantics. 
To achieve this, we design a semantic and contour detector that adopts a cascade structure of cross-attention and self-attention. 
This approach uses global semantic information garnered from self-attention to autonomously select out local details situated at the borders of the contours. 
On this base, we further utilize the MLP-based extraction branch to purify contour details, thereby significantly enhancing the effect of refinement.

% refinement effect

% \begin{figure}
% \centering
% \includegraphics[width=110mm]{our_network.png} 
% \caption{
% The Architecture of the EFormer. It contains backone for encoder, transformer block for decoder and the prediction stage, where transformer block includes a semantic and contour detector, 
% a semantic extraction branch, and a contour-edge extraction branch.}
% \label{fig:network}
% \end{figure}
\begin{figure*}
  \centering
  % \begin{subfigure}{0.68\linewidth}
  %   \fbox{\rule{0pt}{2in} \rule{.9\linewidth}{0pt}}
  %   \caption{An example of a subfigure.}
  %   \label{fig:short-a}
  % \end{subfigure}
  % \hfill
  % \begin{subfigure}{0.28\linewidth}
  %   \fbox{\rule{0pt}{2in} \rule{.9\linewidth}{0pt}}
  %   \caption{Another example of a subfigure.}
  %   \label{fig:short-b}
  % \end{subfigure}
  \includegraphics[width=1 \linewidth]{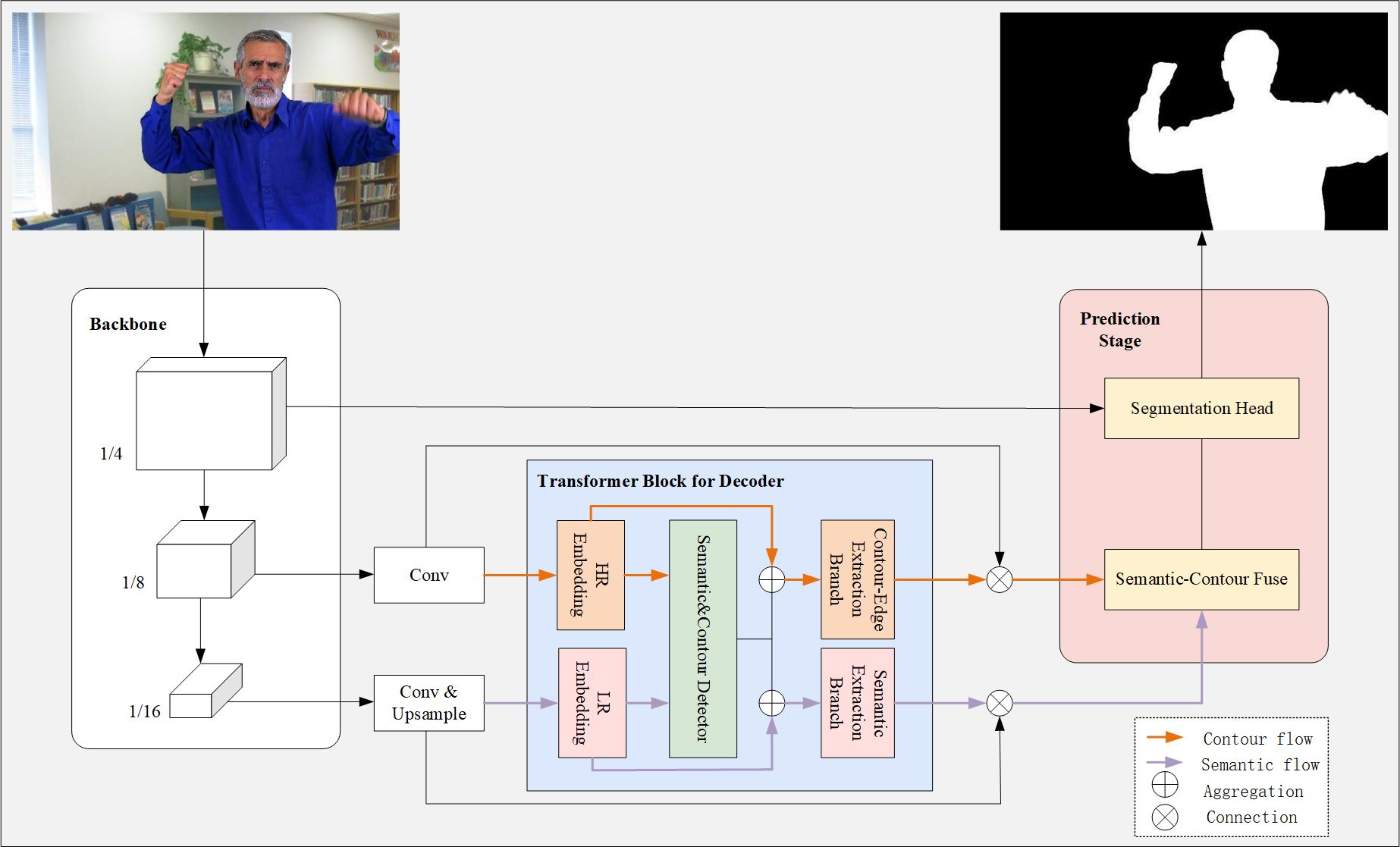}
  \caption{
  The Architecture of the EFormer. It contains backone for encoder, transformer block for decoder and the prediction stage, where transformer block includes a semantic and contour detector, a semantic extraction branch, and a contour-edge extraction branch.}
  \label{fig:network}
\end{figure*}

%% file: sec/3_Model.tex
\section{Model}
\label{sec:Related Works}
In this section, we introduce the process of establishing the complete model architecture of EFormer, which comprises backbone for encoder, transformer block for decoder and the prediction stage, as shown in \cref{fig:network}.
%------------------------------------------------------------------------- 
\subsection{Backbone for Encoder}
The CNN backbone takes RGB images $I\in\mathbb{R}^{B\times 3\times H\times W}$ as input, where $B$ is the number of images, to obtain a pyramid of feature maps $F=\left\{F_{enc\frac{1}{4}},F_{enc\frac{1}{8}},F_{enc\frac{1}{16}}\right\}$ with dimension $F_{enc\frac{1}{i}}\in\mathbb{R}^{B\times C_i\times \frac{H}{i}\times \frac{W}{i}}$.
As we all known, in the pyramid of feature maps, shallow features have high resolution and rich details, while deep features have low resolution but more concentrated semantic information.
Based on this, our research makes the cross-attention mechanism to filter the high-frequency contour features from high-resolution features, using low-frequency semantic information in low-resolution features.
However, since the large resolution of feature maps would bring large computation cost, we use the $F_{enc\frac{1}{8}}$ as $F_{HR}$ (high-resolution features) and use the $F_{enc\frac{1}{16}}$ as $F_{LR}$ (low-resolution features) in our base model.
% the $F_{enc\frac{1}{8}}$ represents high-resolution features as $F_{HR}$ and the $F_{enc\frac{1}{16}}$ represents low-resolution features as $F_{LR}$ in our base model.
The impact of other choices of $F_{HR}$ and $F_{LR}$ on model performance is demonstrated in ablation experiments \cref{Ablation Studies}.
As shown in \cref{fig:network}, the $F_{HR}$ serves as the source of contour flow, providing high-frequency contour detail features for the following processes. Correspondingly, the $F_{LR}$ serves as the source of semantic flow, providing low-frequency semantic information.
Then, the $F_{HR}$ and $F_{LR}$ are projected to same dimension $C$ and respectively flattened into high-resolution feature embedding as $F_{HR}^{em}$ and low-resolution feature embedding as $F_{LR}^{em}$, where $F_{HR}^{em}\in R^{N\times B\times C}$, $F_{LR}^{em}\in R^{N\times B\times C}$, and $N=\frac{H}{8}\times \frac{W}{8}$.
Both of them are utilized as inputs for the transformer block.
%------------------------------------------------------------------------- 
\subsection{Transformer Block for Decoder}
\begin{figure}[t]
  \centering
  % \fbox{\rule{0pt}{2in} \rule{0.9\linewidth}{0pt}}
   \includegraphics[width=0.9\linewidth]{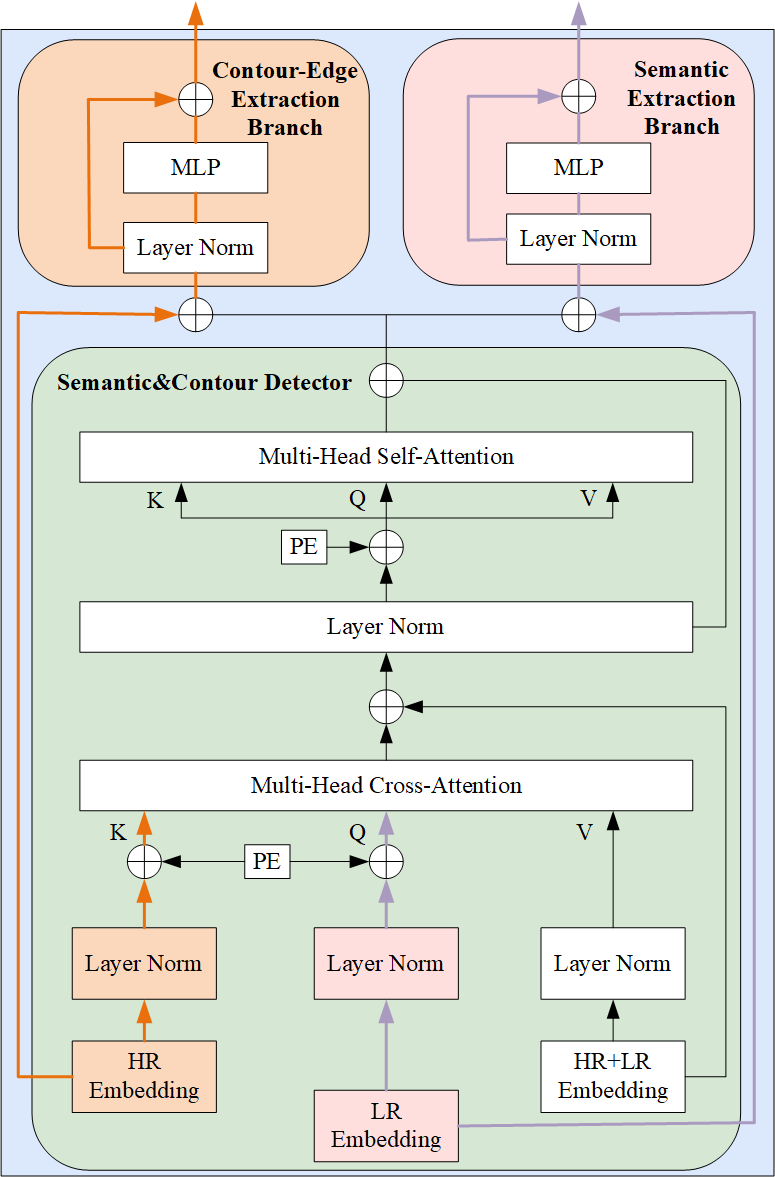}
   \caption{
    Our transformer block. It includes a Semantic and Contour Detector (SCD), a Semantic Extraction Branch (SEB), and a Contour-Edge Extraction Branch (CEEB). The detailed composition of the module is shown in the figure.}
   \label{fig:Transformer Block for Decoder}
\end{figure}

We propose a novel transformer block in the model decoding stage to enhance the ability to capture and extract visual features. 
It not only does not reduce the focus on low-frequency semantic features, but also significantly increases the attention to high-frequency contour features. 
Its detailed architecture is depicted in \cref{fig:Transformer Block for Decoder}. 
The novel transformer block firstly uses a Semantic and Contour Detector (SCD) to locate and capture high-frequency contour features and low-frequency semantic features. 
To achieve this, we cascade a cross-attention layer (CA) and a self-attention layer (SA) in the SCD.
Then, as shown in \cref{fig:Transformer Block for Decoder}, a Contour-Edge Extraction Branch (CEEB) and a Semantic Extraction Branch (SEB) are further built to handle high-frequency contour feature flow and low-frequency semantic feature flow respectively. 
These branches can separately purify more refined high-frequency contour features and extract more complete low-frequency semantic features.
This ensures an outcome that is optimal for both types of features. 
% \begin{figure}
% \centering
% \includegraphics[width=120mm]{our_transformer_block.png} 
% \caption{
% Our transformer block. It includes a Semantic and Contour Detector (SCD), a Semantic Extraction Branch (SEB), and a Contour-Edge Extraction Branch (CEEB). The detailed composition of the module is shown in the figure.}
% \label{fig:Transformer Block for Decoder}
% \end{figure}

% %------------------------------------------------------------------------- 
% \subsubsection{Semantic and Contour Detector (SCD)}
% \textbf{Semantic and Contour Detector (SCD)}
\noindent
{\bf Semantic and Contour Detector (SCD).}
There is a fact that low-frequency semantic features far outperform high-frequency contour detail features in terms of quantity and distribution density.
So that, in the global information propagation process, the low-frequency semantic information easily dominates the feature representations of the original transformer block with self-attention module.

To prevent the loss of high-frequency contour detail information, we first use a cross-attention layer with different resolution features rather than self-attention layer. 
The cross-attention layer can capture high-frequency detail features with high semantic correlation in high-resolution features, based on the low-frequency semantic information in low-resolution features.
Technically, after getting $F_{HR}^{em}$ and $F_{LR}^{em}$, we use Layer Norm (LN) processing them to maintain consistency.
Due to the sensitivity of attention mechanism towards positional information, we use absolute and learnable position encoding (PE) in the SCD.
The input $K$, $Q$, $V$ of the cross-attention layer (CA) is calculated as
\begin{equation}
  K=\ LN\left(F_{HR}^{em}\right)+PE
  \label{equ:dt}
\end{equation}
\begin{equation}
  Q=\ LN\left(F_{LR}^{em}\right)+PE
  \label{equ:dt}
\end{equation}
\begin{equation}
  V=\ LN\left(F_{LR}^{em}+\ F_{HR}^{em}\right)
  \label{equ:dt}
\end{equation}
By performing cross-attention calculations using features of different resolutions, it is possible to automatically locate and capture  the discrete contours along the portrait and allocate more attention to corresponding positions. The contour-edge features can be derived from the output of the CA.
\begin{equation}
  F_{contour-edge}=CA\left(K,\ Q,\ V\right)
  \label{equ:dt}
\end{equation}
Although CA pays extra attention to contour-edge features, it still captures the overall semantic. By the residual connection between CA and $V$, both the semantic and contour-edge features of the portrait are enhanced, especially the contour-edge features.
\begin{equation}
  F_{enhance}=CA\left(K,\ Q,\ V\right)+V
  \label{equ:dt}
\end{equation}
Since semantic information is still present, we use the self-attention layer (SA) after CA.
SA can locate the semantic feature within the portrait and contours, thereby emphasize the semantic attributes at their respective positions.
After using LN to normalize $F_{enhance}$, the input $K^\prime$, $Q^\prime$, $V^\prime$ of the SA is calculated as
\begin{equation}
  K^\prime=Q^\prime=LN\left(F_{enhance}\right)+PE
  \label{equ:dt}
\end{equation}
\begin{equation}
  V^\prime=LN\left(F_{enhance}\right)
  \label{equ:dt}
\end{equation}
\begin{equation}
  F_{semantic-contour}=SA\left(K^\prime,\ Q^\prime,\ V^\prime\right)+V^\prime
  \label{equ:dt}
\end{equation}
Through SA filtering the semantic and contour features, residual connection between $V^\prime$ and the output of SA, we can produce $F_{semantic-contour}$ as the output of the semantic and contour detector. 

% %------------------------------------------------------------------------- 
% \subsubsection{Feature Extraction Branches}
\noindent
{\bf Feature Extraction Branches.}
After SCD, we use multi-layer perceptron (MLP) to build a Contour-Edge Extraction Branch (CEEB) and a Semantic Extraction Branch (SEB).
To prevent the loss of high-frequency contour detail information, we aggregate $F_{semantic-contour}$ getting from SCD and $F_{HR}^{em}$ in the contour flow.
After the merged feature normalized by LN, We employ CEEB to purify it, in order to further extract and refine high-frequency contour details individually.
Similarly, to obtain more comprehensive and consistent portrait semantic features separately, we merge $F_{semantic-contour}$ and $F_{LR}^{em}$ in the semantic flow.
Then, we normalize the aggregated feature applying LN and use SEB to filter it.
\begin{equation}
  F_{contour}=MLP\left(LN\left(F_{HR}^{em}+\ F_{semantic-contour}\right)\right)
  \label{equ:dt}
\end{equation}
\begin{equation}
  F_{semantic}=MLP\left(LN\left(F_{LR}^{em}+\ F_{semantic-contour}\right)\right)
  \label{equ:dt}
\end{equation}
%------------------------------------------------------------------------- 
\subsection{Prediction Stage}
After getting key characteristics $F_{semantic}\in R^{N\times B\times C}$ and $F_{contour}\in R^{N\times B\times C}$ from transformer block, where $N=\frac{H}{8}\times \frac{W}{8}$, we transform them into $F_{semantic}^\prime\in R^{B\times C\times \frac{H}{8}\times \frac{W}{8}}$ and $F_{contour}^\prime\in R^{B\times C\times \frac{H}{8}\times \frac{W}{8}}$ by converting the vector into the matrix. To achieve improved outcomes in portrait segmentation, we propose fusing the two kinds of features and ultimately acquiring the predicted portrait matte from the segmentation head module.

% %------------------------------------------------------------------------- 
% \subsubsection{Fuse}
\noindent
{\bf Fuse.}
We attain the semantic feature through residual connection between $F_{LR}^\prime$ and $F_{semantic}^\prime$ of semantic flow. Similarly, we get contour feature from contour flow and subsequently merge the $F_{semantic}^\prime$ and $F_{contour}^\prime$ to generate the fused features $F_{semantic-contour}^\prime$.
\begin{equation}
  F_{semantic}^\prime=Conv\left(F_{semantic}^\prime+\ F_{LR}^\prime\right)
  \label{equ:dt}
\end{equation}
\begin{equation}
  F_{contour}^\prime=Conv\left(F_{contour}^\prime+\ F_{HR}^\prime\right)
  \label{equ:dt}
\end{equation}
\begin{equation}
  F_{semantic-contour}^\prime=Fuse\left(F_{semantic}^\prime+\ F_{contour}^\prime\right)
  \label{equ:dt}
\end{equation}
% \begin{equation}
%   \begin{split}
%   F_{semantic-contour-fuse}= \\ 
%   Upsample\left(Conv\left(F_{semantic}^\prime+\ F_{contour}^\prime\right)\right)
%   \end{split}
%   \label{equ:dt}
% \end{equation}
% \begin{figure}[t]
%   \centering
%    \begin{equation}
%    F_{semantic-contour-fuse}=
%    Upsample\left(Conv\left(F_{semantic}^\prime+\ F_{contour}^\prime\right)\right)
%    \label{equ:dt}
%    \end{equation}
% \end{figure}

% %------------------------------------------------------------------------- 
% \subsubsection{Segmentation Head}
\noindent
{\bf Segmentation Head.}
Finally, We gather $F_{semantic-contour}^\prime$ and $F_{enc\frac{1}{4}}$ with higher resolution to predict portrait matte in 1/4 scale of the input image and upscale it to the original scale through bi-linear interpolation as the output of our model.
\begin{equation}
  matte=Head\left(F_{semantic-contour}^\prime+F_{enc\frac{1}{4}}\right)
  \label{equ:dt}
\end{equation}

%% file: sec/4_Experiments.tex
\section{Experiments}
\label{sec:Experiments}
%------------------------------------------------------------------------- 
\subsection{Dataset and Evaluation}
% %------------------------------------------------------------------------- 
% \subsubsection{Dataset}
{\bf Dataset.}
We utilize composite training data from two sources: the foreground image dataset VideoMatte240K (JPEG SD Format)\cite{Other-trimap-free-solutions-background-matting-series-VideoMatte240K} and Adobe Image Matting (AIM)\cite{Such-trimap-based-approaches04-AIM}. For the background images, we select BG10K following the approach of BGMv2\cite{Other-trimap-free-solutions-background-matting-series-VideoMatte240K}, which is a collection of photographs depicting various life scenes without any human portraits. We select foregrounds from VideoMatte240K-JPEG-SD and AIM, backgrounds from BG10K to composite image datasets. We split VideoMatte240K-JPEG-SD into 234,982/3,007/2720 image sets with image resolutions of $224 \times 224$ and $512 \times 288$ for training, validating, and testing our model. Similarly, BG10K is split into 9000/1000 image sets, and AIM is split into 214/10 image sets for training and testing, where AIM only includes the portrait images with resolutions at $512 \times 512$. Finally, we use the aforementioned testing datasets to compare and evaluate the performance of EFormer against other models.

% %------------------------------------------------------------------------- 
% \subsubsection{Evaluation}
\noindent
{\bf Evaluation.}
We mainly consider portrait matting accuracy for evaluation. For portrait matting accuracy, we use Mean Absolute Difference (MAD), Mean Squared Error (MSE), Gradient (Grad), and Connectivity (Conn) as evaluation metrics. We also scale MAD, MSE, Grad, and Conn by${10}^{3}$, ${10}^{3}$, ${10}^{-3}$, and ${10}^{-3}$ respectively, for convenience of reference. For all these metrics, the lower number represents better performance.
%------------------------------------------------------------------------- 
\subsection{Implementation Details}
% %------------------------------------------------------------------------- 
% \subsubsection{Training Settings}
{\bf Training Settings.}
For training the network, we use a single RTX 3090 GPU with batch size at 24. The optimizer is AdamW and the initial learning rate is set to ${10}^{-4}$. All models in ablation studies are trained for 25 epochs, with the learning rate decaying by a factor of 0.8 every 5 epochs. Additionally, to augment the data, each input image is subjected to random horizontal flipping.

% %------------------------------------------------------------------------- 
% \subsubsection{Backbone}
\noindent
{\bf Backbone.}
The backbone we use for the network is the pretrained ResNet50\cite{resnet50-101}, since it is the most frequently used backbone in prior works. We use the implementation and weights from torchvision.

% %------------------------------------------------------------------------- 
% \subsubsection{Transformer Block for Decoder}
\noindent
{\bf Transformer Block for Decoder.}
The transformer block is based on the transformer decoder of DETR. We continuously use four transformer blocks in decoder of the network. We set the channel $C=256$ and the number of attention heads $M=8$ for each multi-head attention module.

% %------------------------------------------------------------------------- 
% \subsubsection{Prediction Stage}
\noindent
{\bf Prediction Stage.}
After obtaining the semantic and contour feature maps of the portrait from the stacked transformer blocks, we fuse both of them. Then, the alpha mattes predicted by the segmentation head are upsampled to the original size of the input image with bi-linear interpolation.
% \begin{figure}
% \centering
% \includegraphics[width=110mm]{display_compare.png} 
% \caption{
% Visualization of portrait matte predictions from MODnet, BGMv2, and EFormer under challenging image from the test set of VideoMatte240K-JPEGSD. Our model shows a better ability to distinguish ambiguous foreground contours, as indicated by the red box. Please zoom in to view the detailed information.}
% \label{fig:display-compare}
% \end{figure}
\begin{figure*}
  \centering
  \includegraphics[width=1 \linewidth]{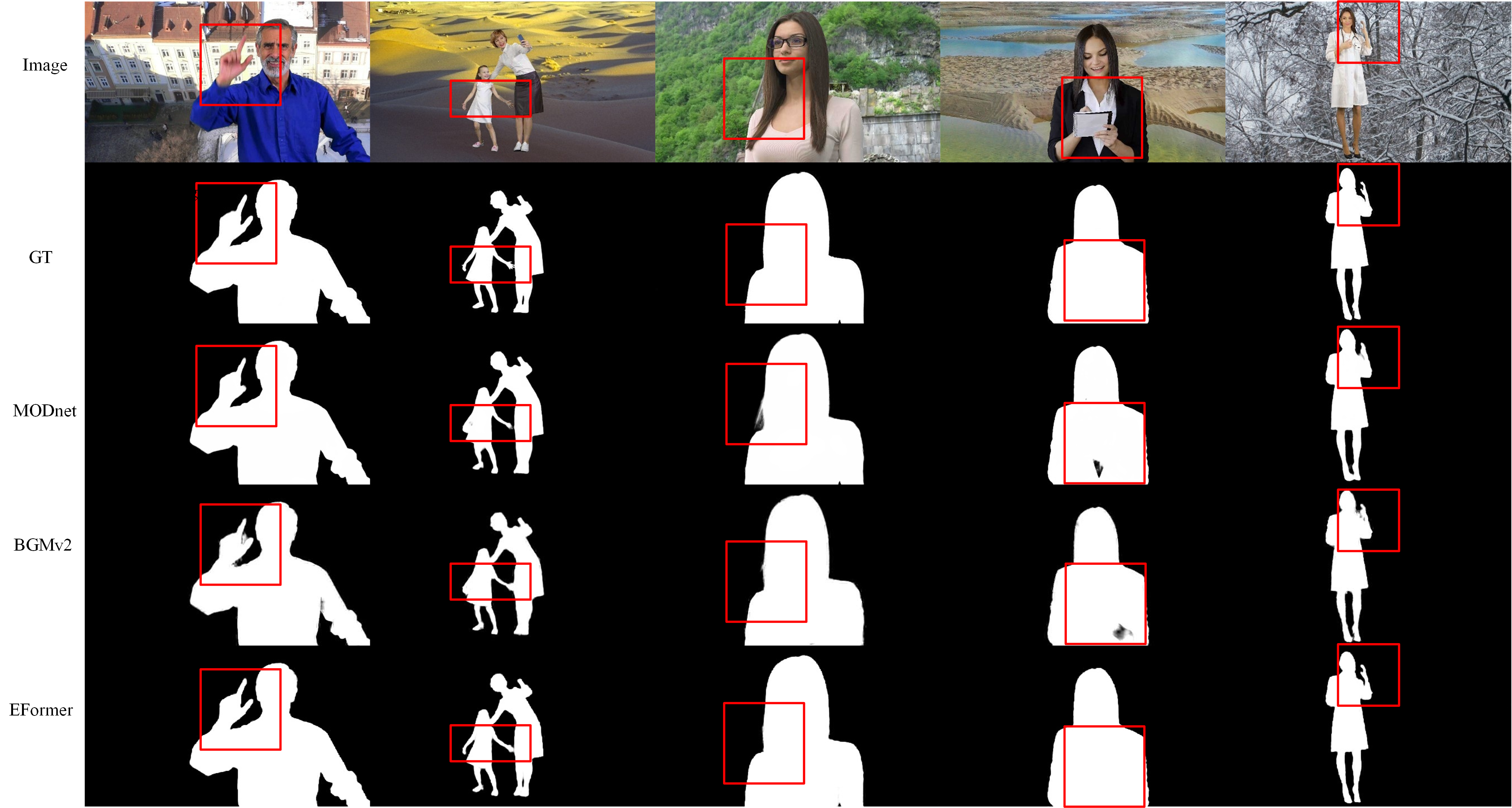}
  \caption{
  Visualization of portrait matte predictions from MODnet, BGMv2, and EFormer under challenging image from the test set of VideoMatte240K (JPEG SD Format). Our model shows a better ability to distinguish ambiguous foreground contours, as indicated by the red box. Please zoom in to view the detailed information.}
  \label{fig:display-compare}
\end{figure*}

% %------------------------------------------------------------------------- 
% \subsubsection{Loss function}
\noindent
{\bf Loss function.}
To compute losses during the training stage, we use a Binary CrossEntropy Loss function.
% \begin{equation}
%   \begin{split}
%   L_{bce}=-\left(g\left(x,y\right)\times\log{\left(p\left(x,y\right)\right)}+ \\ 
%   \left(1-g\left(x,y\right)\right)\times\log\left(1-p\left(x,y\right)\right)\right)
%   \label{equ:dt}
%   \end{split}
% \end{equation}
% Where, $p\left(x,y\right)$ is prediction of network, $g\left(x,y\right)$ is ground truth.
% \begin{equation}
%   L_{bce}=-\left(g\left(x,y\right)\times\log{\left(p\left(x,y\right)\right)}+ \\ 
%   \left(1-g\left(x,y\right)\right)\times\log\left(1-p\left(x,y\right)\right)\right)
%   \label{equ:dt}
% \end{equation}
\begin{equation}
  L_{bce}=-\left(g\times\log{\left(m\right)}+\left(1-g\right)\times\log\left(1-m\right)\right)
  \label{equ:dt}
\end{equation}
Where, $m$ is the prediction of network, $g$ is ground truth.
%
% \setlength{\tabcolsep}{4pt}
% \begin{table}
% \begin{center}
% \caption{
% Comparisons on the test set of VideoMatte240K-JPEG-SD. {\bf Bold} indicates the best performance among these models under the inputs with the same resolution at $512\times 288$.
% }
% \label{table:VM-resolution-512}
% \begin{tabular}{lllll}
% \hline\noalign{\smallskip}
% Model $\qquad$& MAD$\downarrow$ & MSE$\downarrow$ & Grad$\downarrow$ & Conn$\downarrow$\\
% \noalign{\smallskip}
% \hline
% \noalign{\smallskip}
% DeepLabv3\cite{DeepLabv3} & 14.4700 & 9.6700 & 8.5500 & 1.6900\\
% MODnet\cite{Other-trimap-free-solutions-MODNet-PPM-100} & 10.3900 & 5.6500 & 2.0200 & 1.0400\\
% BGMv2\cite{Other-trimap-free-solutions-background-matting-series-VideoMatte240K} & 4.1858 & 1.7934 & 1.3011 & 0.5326\\
% RVM\cite{Other-trimap-free-solutions-RVM} & 5.9900 & 1.1700 &  1.1000 & 0.3400\\
% EFormer & {\bf 2.3097} & {\bf 0.6637} & {\bf 0.4580} & {\bf 0.2740}\\
% \hline
% \end{tabular}
% \end{center}
% \end{table}
% \setlength{\tabcolsep}{1.4pt}
\begin{table}
  \centering
  \begin{tabular}{@{}lllll@{}}
    \toprule
    Model $\qquad$ & MAD$\downarrow$ & MSE$\downarrow$ & Grad$\downarrow$ & Conn$\downarrow$\\
    \midrule
    DeepLabv3\cite{DeepLabv3} & 14.4700 & 9.6700 & 8.5500 & 1.6900\\
    MODnet\cite{Other-trimap-free-solutions-MODNet-PPM-100} & 10.3900 & 5.6500 & 2.0200 & 1.0400\\
    BGMv2\cite{Other-trimap-free-solutions-background-matting-series-VideoMatte240K} & 4.1858 & 1.7934 & 1.3011 & 0.5326\\
    RVM\cite{Other-trimap-free-solutions-RVM} & 5.9900 & 1.1700 &  1.1000 & 0.3400\\
    EFormer & {\bf 2.3097} & {\bf 0.6637} & {\bf 0.4580} & {\bf 0.2740}\\
    \bottomrule
  \end{tabular}
  \caption{
  Comparisons on the test set of VideoMatte240K-JPEG-SD. {\bf Bold} indicates the best performance among these models under the inputs with the same resolution at $512\times 288$.}
  \label{table:VM-resolution-512}
\end{table}
%
% \setlength{\tabcolsep}{4pt}
% \begin{table}
% \begin{center}
% \caption{
% Comparisons on the test set of AIM. {\bf Bold} indicates the best performance among these models under the inputs with the same resolution at $512\times 512$.
% }
% \label{table:AIM-resolution-512}
% \begin{tabular}{lllll}
% \hline\noalign{\smallskip}
% Model $\qquad$& MAD$\downarrow$ & MSE$\downarrow$ & Grad$\downarrow$ & Conn$\downarrow$\\
% \noalign{\smallskip}
% \hline
% \noalign{\smallskip}
% DeepLabv3\cite{DeepLabv3} & 29.64 & 23.78 & 20.17 & 7.71\\
% MODnet\cite{Other-trimap-free-solutions-MODNet-PPM-100} & 21.66 & 14.27 & 5.37 & 5.23\\
% BGMv2\cite{Other-trimap-free-solutions-background-matting-series-VideoMatte240K} & 44.61 & 39.08 & 5.54 & 11.60\\
% RVM\cite{Other-trimap-free-solutions-RVM} & 14.84 & 8.93 &  4.35 & 3.83\\
% EFormer & {\bf 7.47} & {\bf 2.13} & {\bf 2.83} & {\bf 1.90}\\
% \hline
% \end{tabular}
% \end{center}
% \end{table}
% \setlength{\tabcolsep}{1.4pt}

\begin{table}
  \centering
  \begin{tabular}{@{}lllll@{}}
    \toprule
    Model $\qquad$& MAD$\downarrow$ & MSE$\downarrow$ & Grad$\downarrow$ & Conn$\downarrow$\\
    \midrule
    DeepLabv3\cite{DeepLabv3} & 29.64 & 23.78 & 20.17 & 7.71\\
    MODnet\cite{Other-trimap-free-solutions-MODNet-PPM-100} & 21.66 & 14.27 & 5.37 & 5.23\\
    BGMv2\cite{Other-trimap-free-solutions-background-matting-series-VideoMatte240K} & 44.61 & 39.08 & 5.54 & 11.60\\
    RVM\cite{Other-trimap-free-solutions-RVM} & 14.84 & 8.93 &  4.35 & 3.83\\
    EFormer & {\bf 7.47} & {\bf 2.13} & {\bf 2.83} & {\bf 1.90}\\
    \bottomrule
  \end{tabular}
  \caption{
  Comparisons on the test set of AIM. {\bf Bold} indicates the best performance among these models under the inputs with the same resolution at $512\times 512$.}
  \label{table:AIM-resolution-512}
\end{table}
%------------------------------------------------------------------------- 
\subsection{Comparisons to State-of-the-art Methods}
To demonstrate the progressiveness of EFormer, we compare it against the latest state-of-the-art trimap-free portrait matting solutions. 
These are including DeepLabV3\cite{DeepLabv3} with ResNet101\cite{resnet50-101} backbone, BGMv2\cite{Other-trimap-free-solutions-background-matting-series-VideoMatte240K} with MobileNetV2\cite{MobileNetV2} backbone, MODNet\cite{Other-trimap-free-solutions-MODNet-PPM-100}, and RVM\cite{Other-trimap-free-solutions-RVM}. 
Our assessment is based on the testing datasets of VideoMatte240K-JPEG-SD and AIM, respectively. 
As the training files for MODNet are not available, we use its official weights. 
Furthermore, since BGMv2, DeepLabV3, and RVM have already been trained on all datasets, our method can be compared with them in a fair manner.

As shown in the \cref{table:VM-resolution-512}, \cref{table:AIM-resolution-512} and \cref{table:VM-resolution-224} , our model exhibits superior performance in portrait matte tasks. It has a stronger capability to distinguish the contour of a portrait. And it can finely segment edge details (\eg fingers), as exemplified in \cref{fig:display-compare}.
%
% \setlength{\tabcolsep}{4pt}
% \begin{table}
% \begin{center}
% \caption{
% Comparisons on the test set of VideoMatte240K-JPEG-SD. {\bf Bold} indicates the best performance among these models under the inputs with the same resolution at $224\times 224$.
% }
% \label{table:VM-resolution-224}
% \begin{tabular}{lllll}
% \hline\noalign{\smallskip}
% Model $\qquad$& MAD$\downarrow$ & MSE$\downarrow$ & Grad$\downarrow$ & Conn$\downarrow$\\
% \noalign{\smallskip}
% \hline
% \noalign{\smallskip}
% DeepLabv3\cite{DeepLabv3} & 18.1586 & 14.7623 & 6.0015 & 2.8824\\
% MODnet\cite{Other-trimap-free-solutions-MODNet-PPM-100} & 13.1400 & 8.9656 & 2.6028 & 2.1543\\
% BGMv2\cite{Other-trimap-free-solutions-background-matting-series-VideoMatte240K} & 7.0471 & 2.3543 & 1.5621 & 1.0651\\
% RVM\cite{Other-trimap-free-solutions-RVM} & 6.1357 & 2.2647 &  0.6523 & 0.2866\\
% EFormer & {\bf 4.1357} & {\bf 1.4280} & {\bf 0.3760} & {\bf 0.1753}\\
% \hline
% \end{tabular}
% \end{center}
% \end{table}
% \setlength{\tabcolsep}{1.4pt}
\begin{table}
  \centering
  \begin{tabular}{@{}lllll@{}}
    \toprule
    Model $\qquad$& MAD$\downarrow$ & MSE$\downarrow$ & Grad$\downarrow$ & Conn$\downarrow$\\
    \midrule
    DeepLabv3\cite{DeepLabv3} & 18.1586 & 14.7623 & 6.0015 & 2.8824\\
    MODnet\cite{Other-trimap-free-solutions-MODNet-PPM-100} & 13.1400 & 8.9656 & 2.6028 & 2.1543\\
    BGMv2\cite{Other-trimap-free-solutions-background-matting-series-VideoMatte240K} & 7.0471 & 2.3543 & 1.5621 & 1.0651\\
    RVM\cite{Other-trimap-free-solutions-RVM} & 6.1357 & 2.2647 &  0.6523 & 0.2866\\
    EFormer & {\bf 4.1357} & {\bf 1.4280} & {\bf 0.3760} & {\bf 0.1753}\\
    \bottomrule
  \end{tabular}
  \caption{
  Comparisons on the test set of VideoMatte240K-JPEG-SD. {\bf Bold} indicates the best performance among these models under the inputs with the same resolution at $224\times 224$.}
  \label{table:VM-resolution-224}
\end{table}
%------------------------------------------------------------------------- 
\subsection{Ablation Studies}
\label{Ablation Studies}
% %------------------------------------------------------------------------- 
% \subsubsection{Training Epochs}
{\bf Training Epochs.}
In the ablation study with an image resolution of $224\times 224$, we conduct 25 training epochs. Every 5 epochs of training, we evaluate the training effect of the model. At the same time, the learning rate is reduced by 0.8 times. The experimental results indicate that our model achieve optimal performance in the 20 epoch, as shown in the \cref{table:Training Epochs}.
%
% \setlength{\tabcolsep}{4pt}
% \begin{table}
% \begin{center}
% \caption{
% Ablation Study on the Training Epochs with the test set of VideoMatte240K-JPEG-SD (resolution at $224\times 224$). {\bf Bold} indicates the best performance among these models. We use the result of 20-epoch training as a strong baseline model.
% }
% \label{table:Training Epochs}
% \begin{tabular}{lllll}
% \hline\noalign{\smallskip}
%  $\qquad$& MAD$\downarrow$ & MSE$\downarrow$ & Grad$\downarrow$ & Conn$\downarrow$\\
% \noalign{\smallskip}
% \hline
% \noalign{\smallskip}
% 5-epoch & 4.3953 & 1.4824 & 0.4089 & 0.1865\\
% 10-epoch & 4.3108 & 1.4766 & 0.3908 & 0.1834\\
% 15-epoch & 4.2867 & 1.5238 & 0.4087 & 0.1833\\
% 20-epoch & {\bf 4.1357} & {\bf 1.4280} & {\bf 0.3760} & {\bf 0.1753}\\
% 25-epoch & 4.1811 & 1.4797 & 0.3858 &  0.1779\\
% \hline
% \end{tabular}
% \end{center}
% \end{table}
% \setlength{\tabcolsep}{1.4pt}
\begin{table}
  \centering
  \begin{tabular}{@{}lllll@{}}
    \toprule
    $\qquad$& MAD$\downarrow$ & MSE$\downarrow$ & Grad$\downarrow$ & Conn$\downarrow$\\
    \midrule
    5-epoch & 4.3953 & 1.4824 & 0.4089 & 0.1865\\
    10-epoch & 4.3108 & 1.4766 & 0.3908 & 0.1834\\
    15-epoch & 4.2867 & 1.5238 & 0.4087 & 0.1833\\
    20-epoch & {\bf 4.1357} & {\bf 1.4280} & {\bf 0.3760} & {\bf 0.1753}\\
    25-epoch & 4.1811 & 1.4797 & 0.3858 &  0.1779\\
    \bottomrule
  \end{tabular}
  \caption{
  Ablation Study on the Training Epochs with the test set of VideoMatte240K-JPEG-SD (resolution at $224\times 224$). {\bf Bold} indicates the best performance among these models. We use the result of 20-epoch training as a strong baseline model.}
  \label{table:Training Epochs}
\end{table}

\begin{figure}[t]
  \centering
  % \fbox{\rule{0pt}{2in} \rule{0.9\linewidth}{0pt}}
   \includegraphics[width=1 \linewidth]{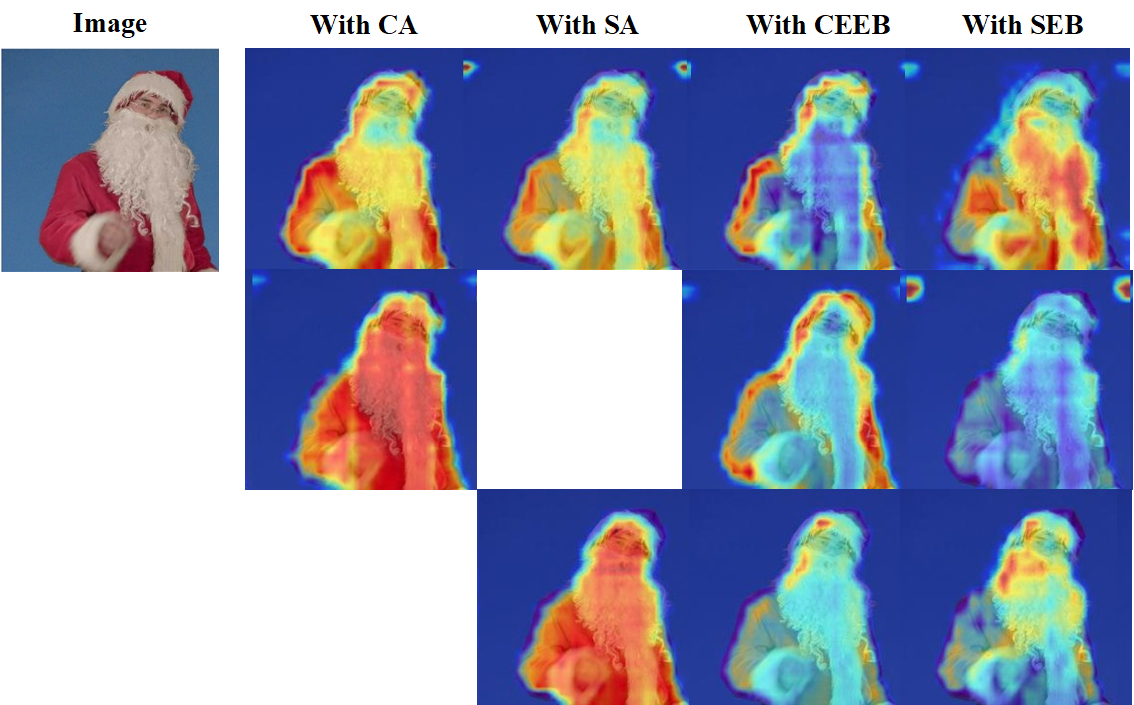}
   \caption{
    We use Grad-CAM to visualize the attention distribution of various features in CA, SA, CEEB, and SEB. The model in the first row contains four complete modules. For comparison, the SA layer and CA layer were removed from the models in the second and third rows, respectively.}
   \label{fig:The Functional Role of Each Module}
\end{figure}

% %------------------------------------------------------------------------- 
% \subsubsection{The Functional Role of Each Module}
\noindent
{\bf The Functional Role of Each Module.}
To demonstrate the effectiveness of the model in capturing and extracting contour details and portrait semantics, we analyze the outputs of CA, SA, CEEB, and SEB, and utilize Class Activation Map (CAM) to visualize the attention allocation of each layer's features, as shown in \cref{fig:The Functional Role of Each Module}.

From the first line of \cref{fig:The Functional Role of Each Module}, it can be evident that after the cross-attention layer, the model clearly strengthens its attention to the contours of the portrait. Following the contour prior, self-attention layer captures the portrait's semantics thoroughly. CEEB and SEB models provide sharper contour and more comprehensive semantic information.

Comparing the second and first lines of \cref{fig:The Functional Role of Each Module}, it is observed that removing the self-attention layer considerably weakens the semantic information extracted by SEB.

In the comparison between the third and first lines of \cref{fig:The Functional Role of Each Module}, it can be indicated that removing the cross-attention layer significantly weakens the contour information extracted by CEEB.

% \begin{figure}
% \centering
% \includegraphics[width=110mm]{ablation_experiment_compare.png} 
% \caption{
% We use Grad-CAM to visualize the attention distribution of various features in CA, SA, CEEB, and SEB. The model in the first row contains four complete modules. For comparison, the SA layer and CA layer were removed from the models in the second and third rows, respectively.}
% \label{fig:The Functional Role of Each Module}
% \end{figure}

After qualitative analysis, the prediction results of all the above models are evaluated. It is concluding that the model with CA, SA, CEEB, and SEB is both effective and rational, as demonstrated in the \cref{table:The Functional Role of Each Module}.

%
% \setlength{\tabcolsep}{4pt}
% \begin{table}
% \begin{center}
% \caption{
% Ablation Study on the Functional Role of CA and SA with the test set of VideoMatte240K-JPEG-SD (resolution at $224\times 224$). {\bf Bold} indicates the best performance among these models.
% }
% \label{table:The Functional Role of Each Module}
% \begin{tabular}{llllll}
% \hline\noalign{\smallskip}
% CA & SA $\qquad$& MAD$\downarrow$ & MSE$\downarrow$ & Grad$\downarrow$ & Conn$\downarrow$\\
% \noalign{\smallskip}
% \hline
% \noalign{\smallskip}
% \checkmark & & 4.2762 & 1.5344 & 0.4092 & 0.1823\\
%  & \checkmark & 4.1710 & 1.4354 & 0.3773 & 0.1769\\
% \checkmark & \checkmark & {\bf 4.1357} & {\bf 1.4280} & {\bf 0.3760} & {\bf 0.1753}\\
% \hline
% \end{tabular}
% \end{center}
% \end{table}
% \setlength{\tabcolsep}{1.4pt}

\begin{table}
  \centering
  \begin{tabular}{@{}llllll@{}}
    \toprule
    CA & SA $\qquad$ & MAD$\downarrow$ & MSE$\downarrow$ & Grad$\downarrow$ & Conn$\downarrow$\\
    \midrule
    \checkmark & & 4.2762 & 1.5344 & 0.4092 & 0.1823\\
     & \checkmark & 4.1710 & 1.4354 & 0.3773 & 0.1769\\
    \checkmark & \checkmark & {\bf 4.1357} & {\bf 1.4280} & {\bf 0.3760} & {\bf 0.1753}\\
    \bottomrule
  \end{tabular}
  \caption{
  Ablation Study on the Functional Role of CA and SA with the test set of VideoMatte240K-JPEG-SD (resolution at $224\times 224$). {\bf Bold} indicates the best performance among these models.}
  \label{table:The Functional Role of Each Module}
\end{table}

% the absence of SA leads to
As is well known, when it comes to extracting features for portrait segmentation tasks, semantic information plays a significant role, in contrast, contour details account for a relatively small proportion. 
Owing to this, as shown in the \cref{table:The Functional Role of Each Module} and \cref{fig:The Functional Role of Each Module}, the model without SA leads to a significant increase in all metrics due to the lack of amount of semantic information, while the model without CA experiences a slight increase in all indicators due to the loss of contour details. 
In order to fully harness the potential of the vision transformer, we integrate both CA and SA to ensure semantic completeness and enhance attention towards contour details, thereby achieving more comprehensive and fine-grained segmentation results.

\noindent
% {\bf The Selection of $F_{HR}$ and $F_{LR}$}
\textbf{The Selection of $F_{HR}$ and $F_{LR}$.}
% As we all known, in the pyramid of feature maps, shallow features have high resolution and rich details, while deep features have low resolution but more concentrated semantic information.
% Based on this, our research make the cross attention mechanism to filter the high-frequency contour features from high-resolution features, using low-frequency semantic information in low-resolution features.
% In our work, the key point is using the cross-attention module between the different resolution features to filter contour feature
In our base model, considering the large resolution of feature maps would bring large computation cost, we use the $F_{enc\frac{1}{8}}$ as $F_{HR}$ (high-resolution features) and use the $F_{enc\frac{1}{16}}$ as $F_{LR}$ (low-resolution features).
Then the $F_{HR}$ serves as the source of contour flow, providing high-frequency contour detail features. 
Correspondingly, the $F_{LR}$ serves as the source of semantic flow, providing low-frequency semantic information.
As shown in the \cref{table:The Source of FHR and FLR}, after using $F_{enc\frac{1}{4}}$ with higher resolution as $F_{HR}$, our model can capture more high-frequency contour detail features and its performance is better.
And after using $F_{enc\frac{1}{16}}$ with lower resolution as $F_{LR}$, our model can use more concentrated low-frequency semantic information, which makes all indicators optimal.
%
% \setlength{\tabcolsep}{4pt}
% \begin{table}
% \begin{center}
% \caption{
% Ablation Study on the The Selection of $F_{HR}$ and $F_{LR}$ with the test set of VideoMatte240K-JPEG-SD (resolution at $224\times 224$). {\bf Bold} indicates the best performance among these models.
% }
% \label{table:The Source of FHR and FLR}
% \begin{tabular}{llllll}
% \hline\noalign{\smallskip}
% $F_{HR}$ & $F_{LR}$ $\qquad$& MAD$\downarrow$ & MSE$\downarrow$ & Grad$\downarrow$ & Conn$\downarrow$\\
% \noalign{\smallskip}
% \hline
% \noalign{\smallskip}
% $F_{enc\frac{1}{8}}$ & $F_{enc\frac{1}{16}}$ & 4.1357 & 1.4280 & 0.3760 & 0.1753\\
% $F_{enc\frac{1}{4}}$ & $F_{enc\frac{1}{8}}$ & 3.9970 & 1.3343 & 0.3328 & 0.1679\\
% $F_{enc\frac{1}{4}}$ & $F_{enc\frac{1}{16}}$ & {\bf 3.9730} & {\bf 1.2949} & {\bf 0.3215} & {\bf 0.1665}\\
% \hline
% \end{tabular}
% \end{center}
% \end{table}
% \setlength{\tabcolsep}{1.4pt}

\begin{table}
  \centering
  \begin{tabular}{@{}llllll@{}}
    \toprule
    $F_{HR}$ & $F_{LR}$ $\qquad$& MAD$\downarrow$ & MSE$\downarrow$ & Grad$\downarrow$ & Conn$\downarrow$\\
    \midrule
    $F_{enc\frac{1}{8}}$ & $F_{enc\frac{1}{16}}$ & 4.1357 & 1.4280 & 0.3760 & 0.1753\\
    $F_{enc\frac{1}{4}}$ & $F_{enc\frac{1}{8}}$ & 3.9970 & 1.3343 & 0.3328 & 0.1679\\
    $F_{enc\frac{1}{4}}$ & $F_{enc\frac{1}{16}}$ & {\bf 3.9730} & {\bf 1.2949} & {\bf 0.3215} & {\bf 0.1665}\\
    \bottomrule
  \end{tabular}
  \caption{
  Ablation Study on the the Selection of $F_{HR}$ and $F_{LR}$ with the test set of VideoMatte240K-JPEG-SD (resolution at $224\times 224$). {\bf Bold} indicates the best performance among these models.}
  \label{table:The Source of FHR and FLR}
\end{table}

%% file: sec/5_Conclusion.tex
\section{Conclusion}
\label{sec:Conclusion}
In this paper, we propose EFormer, which can strengthen the transformer's attention to semantic and contour features of foreground, improving the prediction accuracy of portrait matting. Our transformer block incorporates cross-attention and self-attention to establish a semantic and contour detector, which successfully guides the model to focus on both low-frequency semantic and high-frequency contours features, especially the contour features. Additionally, we further build a contour-edge extraction branch and a semantic extraction branch to extract finer high-frequency contour details and more comprehensive portrait's low-frequency semantic information. Extensive experiments show that our method outperforms previous portrait matting solutions and exhibits robust performance during testing and inference.

% To deal with this issue, we propose EFormer to enhance the model's attention towards both of the low-frequency semantic and high-frequency contour features. 
% For the high-frequency contours, our research demonstrates that cross-attention module between different resolutions can guide our model to allocate attention appropriately to these contour regions.
% Supported on this, we can successfully extract the high-frequency detail information around the portrait's contours, which are previously ignored by self-attention.
% Based on cross-attention module, we further build a semantic and contour detector (SCD) to accurately capture both of the low-frequency semantic and high-frequency contour features.
% And we design contour-edge extraction branch and semantic extraction branch to extract refined high-frequency contour features and complete low-frequency semantic information, respectively.